# CNN-RNN: A Unified Framework for Multi-label Image Classification

Jiang Wang[1]   Yi Yang[1]   Junhua Mao[2]   Zhiheng Huang[3*]   Chang Huang[4*]   Wei Xu[1]
[1]Baidu Research   [2]University of California at Los Angles   [3]Facebook Speech   [4] Horizon Robotics

## Abstract

*While deep convolutional neural networks (CNNs) have shown a great success in single-label image classification, it is important to note that real world images generally contain multiple labels, which could correspond to different objects, scenes, actions and attributes in an image. Traditional approaches to multi-label image classification learn independent classifiers for each category and employ ranking or thresholding on the classification results. These techniques, although working well, fail to explicitly exploit the label dependencies in an image. In this paper, we utilize recurrent neural networks (RNNs) to address this problem. Combined with CNNs, the proposed CNN-RNN framework learns a joint image-label embedding to characterize the semantic label dependency as well as the image-label relevance, and it can be trained end-to-end from scratch to integrate both information in a unified framework. Experimental results on public benchmark datasets demonstrate that the proposed architecture achieves better performance than the state-of-the-art multi-label classification models.*

## 1. Introduction

Every real-world image can be annotated with multiple labels, because an image normally abounds with rich semantic information, such as objects, parts, scenes, actions, and their interactions or attributes. Modeling the rich semantic information and their dependencies is essential for image understanding. As a result, *multi-label* classification task is receiving increasing attention [12, 9, 24, 36]. Inspired by the great success from deep convolutional neural networks in *single-label* image classification in the past few years [17, 29, 32], which demonstrates the effectiveness of end-to-end frameworks, we explore to learn a unified framework for *multi-label* image classification.

A common approach that extends CNNs to multi-label classification is to transform it into multiple single-label classification problems, which can be trained with the ranking loss [9] or the cross-entropy loss [12]. However, when treating labels independently, these methods fail to model

---

[*]This work was done when the authors are at Baidu Research.

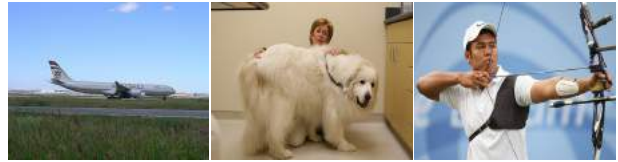

| Airplane | Great Pyrenees | Archery |

*Sky, Grass, Runway   Dog, Person, Room   Person, Hat, Nike*

Figure 1. We show three images randomly selected from ImageNet 2012 classification dataset. The second row shows their corresponding label annotations. For each image, there is only one label (*i.e.* Airplane, Great Pyrenees, Archery) annotated in the ImageNet dataset. However, every image actually contains *multiple labels*, as suggested in the third row.

the dependency between multiple labels. Previous works have shown that multi-label classification problems exhibit strong label co-occurrence dependencies [39]. For instance, *sky* and *cloud* usually appear together, while *water* and *cars* almost never co-occur.

To model label dependency, most existing works are based on graphical models [39], among which a common approach is to model the co-occurrence dependencies with pairwise compatibility probabilities or co-occurrence probabilities and use Markov random fields [13] to infer the final joint label probability. However, when dealing with a large set of labels, the parameters of these pairwise probabilities can be prohibitively large while lots of the parameters are redundant if the labels have highly overlapping meanings. Moreover, most of these methods either can not model higher-order correlations [39], or sacrifice computational complexity to model more complicated label relationships [20]. In this paper, we explicitly model the label dependencies with recurrent neural networks (RNNs) to capture higher-order label relationships while keeping the computational complexity tractable. We find that RNN significantly improves classification accuracy.

For the CNN part, to avoid problems like overfitting, previous methods normally assume all classifiers share the same image features [36]. However, when using the same image features to predict multiple labels, objects that are small in the images are easily get ignored or hard to recognize independently. In this work, we design the RNNs framework to adapt the image features based on the previous prediction results, by encoding the attention models implicitly in the CNN-RNN structure. The idea behind it is to implicitly adapt the attentional area in images so the CNNs can focus its attention on different regions of the images when predicting different labels. For example, when predicting multiple labels for images in Figure 1, our model will shift its attention to smaller ones (*i.e.* Runway, Person, Hat) after recognizing the dominant object (*i.e.* Airplane, Great Pyrenees, Archery). These small objects are hard to recognize by itself, but can be easily inferred given enough contexts.

Finally, many image labels have overlapping meanings. For example, *cat* and *kitten* have almost the same meanings and are often interchangeable. Not only does exploiting the semantic redundancies reduce the computational cost, it also improves the generalization ability because the labels with duplicate semantics can get more training data.

The label semantic redundancy can be exploited by joint image/label embedding, which can be learned via canonical correlation analysis [10], metric learning [19], or learning to rank methods [37]. The joint image/label embedding maps each label or image to an embedding vector in a joint low-dimensional Euclidean space such that the embeddings of semantically similar labels are close to each other, and the embedding of each image should be close to that of its associated labels in the same space. The joint embedding model can exploit label semantic redundancy because it essentially shares classification parameters for semantically similar labels. However, the label co-occurrence dependency is largely ignored in most of these models.

In this paper, we propose a unified CNN-RNN framework for multi-label image classification, which effectively learns both the semantic redundancy and the co-occurrence dependency in an end-to-end way. The framework of the proposed model is shown in Figure 2. The multi-label RNN model learns a joint low-dimensional image-label embedding to model the semantic relevance between images and labels. The image embedding vectors are generated by a deep CNN while each label has its own label embedding vector. The high-order label co-occurrence dependency in this low-dimensional space is modeled with the long short term memory recurrent neurons, which maintains the information of label context in their internal memory states. The RNN framework computes the probability of a multi-label prediction sequentially as an ordered prediction path, where the a priori probability of a label at each time step can be computed based on the image embedding and the output of the recurrent neurons. During prediction, the multi-label prediction with the highest probability can be approximately found with beam search algorithm. The proposed CNN-RNN framework is a unified framework which combines the advantages of the joint image/label embedding

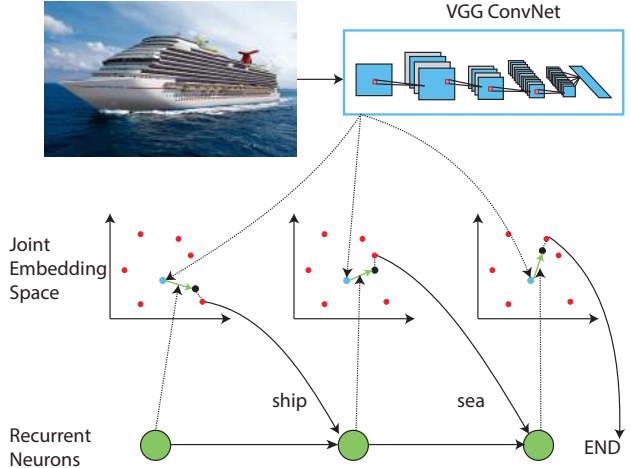

Figure 2. An illustration of the CNN-RNN framework for multi-label image classification. The framework learns a joint embedding space to characterize the image-label relationship as well as label dependency. The red and blue dots are the label and image embeddings, respectively, and the black dots are the sum of the image and recurrent neuron output embeddings. The recurrent neurons model the label co-occurrence dependencies in the joint embedding space by sequentially linking the label embeddings in the joint embedding space. At each time step, the probability of a label is computed based on the image embedding and the output of the recurrent neurons. (best viewed in color)

and label co-occurrence models, and it can be trained in an end-to-end way.

Compared with state-of-the-art multi-label image classification methods, the proposed RNN framework has several advantages:

- The framework employs an end-to-end model to utilize the semantic redundancy and the co-occurrence dependency, both of which is indispensable for effective multi-label classifications.

- The recurrent neurons is more compact and more powerful model of high-order label co-occurrence dependency than other label co-occurrence models in this task.

- The implicit attention mechanism in the recurrent neurons adapt the image features to better predict small objects that need more contexts.

We evaluate the proposed CNN-RNN framework with exhaustive experiments on public multi-label benchmark datasets inlcuding NUS-WIDE, Microsoft COCO, and PASCAL VOC 2007. Experimental results demonstrate that the proposed method achieves significantly better performance compared to the current state-of-the-art multi-label classification methods. We also visualize the attentional regions of the RNN framework with the Deconvolutional net-works [40]. Interestingly, the visualization shows that the RNN framework can focus on the corresponding image regions when predicting different labels, which is very similar to humans' multi-label classification process.

## 2. Related Work

The progress of image classification is partly due to the creation of large-scale hand-labeled datasets such as ImageNet [5], and the development of deep convolutional neural networks [17]. Recent work that extends deep convolutional neural networks to multi-label classification achieves good results. Deep convolutional ranking [9] optimizes a top-$k$ ranking objective, which assigns smaller weights to the loss if the positive label. Hypotheses-CNN-Pooling [36] employs max pooling to aggregate the predictions from multiple hypothesis region proposals. These methods largely treat each label independently and ignore the correlations between labels.

Multi-label classification can also be achieved by learning a joint image/label embedding. Multiview Canonical Correlation Analysis [10] is a three-way canonical analysis that maps the image, label, and the semantics into the same latent space. WASABI [37] and DEVISE [7] learn the joint embedding using the learning to rank framework with WARP loss. Metric learning [19] learns a discriminative metric to measure the image/label similarity. Matrix completion [1] and bloom filter [3] can also be employed as label encodings. These methods effectively exploit the label semantic redundancy, but they fall short on modeling the label co-occurrence dependency.

Various approaches have been proposed to exploit the label co-occurrence dependency for multi-label image classification. [28] learns a chain of binary classifiers, where each classifier predicts whether the current label exists given the input feature and the already predicted labels. The label co-occurrence dependency can also be modeled by graphical models, such as Conditional Random Field [8], Dependency Network [13], and co-occurrence matrix[39]. Label augment model [20] augments the label set with common label combinations. Most of these models only capture pairwise label correlations and have high computation cost when the number of labels is large. The low-dimensional recurrent neurons in the proposed RNN model are more computationally efficient representations for high-order label correlation.

RNN with LSTM can effectively model the long-term temporal dependency in a sequence. It has been successfully applied in image captioning [25, 35], machine translation [31], speech recognition [11], language modeling [30], and word embedding learning [18]. We demonstrate that RNN with LSTM is also an effective model for label dependency.

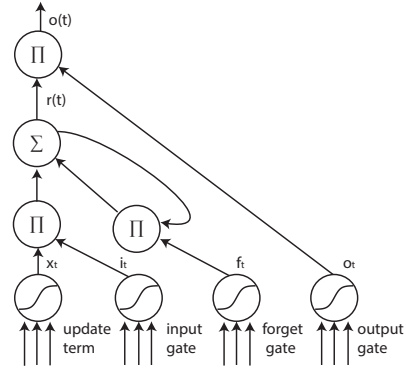

Figure 3. A schematic illustration of a LSTM neuron. Each LSTM neuron has an input gate, a forget gate, and an output gate.

## 3. Method

Since we aim to characterize the high-order label correlation, we employ long short term memory (LSTM) neurons [15] as our recurrent neurons, which has been demonstrated to be a powerful model of long-term dependency.

### 3.1. Long Short Term Memory Networks (LSTM)

RNN [15] is a class of neural network that maintains internal hidden states to model the dynamic temporal behaviour of sequences with arbitrary lengths through directed cyclic connections between its units. It can be considered as a hidden Markov model extension that employs nonlinear transition function and is capable of modeling long term temporal dependencies. LSTM extends RNN by adding three gates to an RNN neuron: a forget gate $f$ to control whether to forget the current state; an input gate $i$ to indicate if it should read the input; an output gate $o$ to control whether to output the state. These gates enable LSTM to learn long-term dependency in a sequence, and make it is easier to optimize, because these gates help the input signal to effectively propagate through the recurrent hidden states $r(t)$ without affecting the output. LSTM also effectively deals with the gradient vanishing/exploding issues that commonly appear during RNN training [26].

$$\begin{aligned}
x_t &= \delta(U_r.r(t-1) + U_w w_k(t)) \\
i_t &= \delta(U_{i_r} r(t-1) + U_{i_w} w_k(t)) \\
f_t &= \delta(U_{f_r} r(t-1) + U_{f_w} w_k(t)) \\
o_t &= \delta(U_{o_r} r(t-1) + U_{o_w} w_k(t)) \\
r(t) &= f_t \odot r(t-1) + i_t \odot x_t \\
o(t) &= r(t) \odot o(t)
\end{aligned} \quad (1)$$

where $\delta(.)$ is an activation function, $\odot$ is the product with gate value, and various $W$ matrices are learned parameters. In our implementation, we employ rectified linear units (ReLU) as the activation function [4].

## 3.2. Model

We propose a novel CNN-RNN framework for multi-label classification problem. The illustration of the CNN-RNN framework is shown in Fig. 4. It contains two parts: The CNN part extracts semantic representations from images; the RNN part models image/label relationship and label dependency.

We decompose a multi-label prediction as an ordered prediction path. For example, labels "zebra" and "elephant" can be decomposed as either ("zebra", "elephant") or ("elephant", "zebra"). The probability of a prediction path can be computed by the RNN network. The image, label, and recurrent representations are projected to the same low-dimensional space to model the image-text relationship as well as the label redundancy. The RNN model is employed as a compact yet powerful representation of the label co-occurrence dependency in this space. It takes the embedding of the predicted label at each time step and maintains a hidden state to model the label co-occurrence information. The a priori probability of a label given the previously predicted labels can be computed according to their dot products with the sum of the image and recurrent embeddings. The probability of a prediction path can be obtained as the product of the a-prior probability of each label given the previous labels in the prediction path.

A label $k$ is represented as a one-hot vector $e_k = [0, \ldots 0, 1, 0, \ldots, 0]$, which is 1 at the $k$-th location, and 0 elsewhere. The label embedding can be obtained by multiplying the one-hot vector with a *label embedding matrix* $U_l$. The $k$-th row of $U_l$ is the label embedding of the label $k$.

$$w_k = U_l.e_k. \quad (2)$$

The dimension of $w_k$ is usually much smaller than the number of labels.

The recurrent layer takes the label embedding of the previously predicted label, and models the co-occurrence dependencies in its hidden *recurrent states* by learning non-linear functions:

$$o(t) = h_o(r(t-1), w_k(t)), \; r(t) = h_r(r(t-1), w_k(t)) \quad (3)$$

where $r(t)$ and $o(t)$ are the hidden states and outputs of the recurrent layer at the time step t, respectively, $w_k(t)$ is the label embedding of the $t$-th label in the prediction path, and $h_o(.), h_r(.)$ are the non-linear RNN functions, which will be described in details in Sec. 3.1.

The output of the recurrent layer and the image representation are projected into the same low-dimensional space as the label embedding.

$$x_t = h(U_o^x o(t) + U_I^x I), \quad (4)$$

where $U_o^x$ and $U_I^x$ are the projection matrices for recurrent layer output and image representation, respectively. The number of columns of $U_o^x$ and $U_I^x$ are the same as the label embedding matrix $U_l$. $I$ is the convolutional neural network image representation. We will show in Sec 4.5 that the learned joint embedding effectively characterizes the relevance of images and labels.

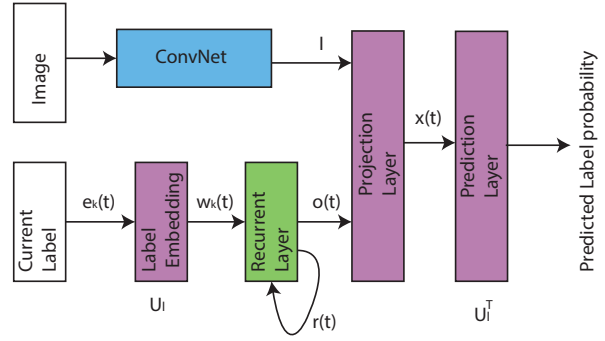

Figure 4. The architecture of the proposed RNN model for multi-label classification. The convolutional neural network is employed as the image representation, and the recurrent layer captures the information of the previously predicted labels. The output label probability is computed according to the image representation and the output of the recurrent layer.

Finally, the label scores can be computed by multiplying the transpose of $U_l$ and $x_t$ to compute the distances between $x_t$ and each label embedding.

$$s(t) = U_l^T x_t. \quad (5)$$

The predicted label probability can be computed using soft-max normalization on the scores.

## 3.3. Inference

A *prediction path* is a sequence of labels $(l_1, l_2, l_3, \cdots, l_N)$, where the probability of each label $l_t$ can be computed with the information of the image $I$ and the previously predicted labels $l_1, \cdots, l_{t-1}$. The RNN model predicts multiple labels by finding the prediction path that maximizes the a priori probability.

$$\begin{aligned} l_1, \cdots, l_k &= \arg \max_{l_1, \cdots, l_k} P(l_1, \cdots, l_k | I) \\ &= \arg \max_{l_1, \cdots, l_k} P(l_1 | I) \times P(l_2 | I, l_1). \quad (6) \\ &\cdots P(l_k | I, l_1, \cdots, l_{k-1}) \end{aligned}$$

Since the probability $P(l_k | I, l_1, \cdots, l_{k-1})$ does not have Markov property, there is no optimal polynomial algorithm to find the optimal prediction path. We can employ the greedy approximation, which predicts label $\hat{l}_t = \arg \max_{l_t} P(l_t | I, l_1, \cdots, l_{t-1})$ at time step $t$ and fix the label prediction $\hat{l}_t$ at later predictions. However, the greedy algorithm is problematic because if the first predicted label is wrong, it is very likely that the whole sequence cannot be correctly predicted. Thus, we employ the beam search algorithm to find the top-ranked prediction path.

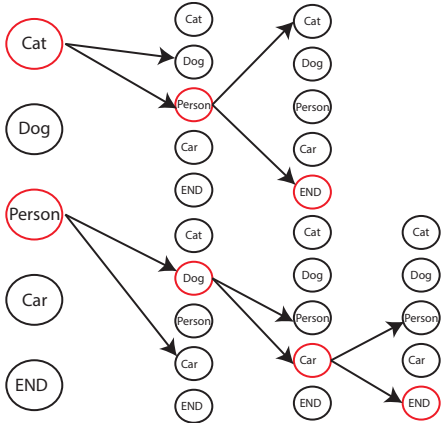

Figure 5. An example of the beam search algorithm with beam size $N = 2$. The beam search algorithm finds the best $N$ paths with the highest probability, by keeping a set of intermediate paths at each time step and iteratively adding labels these intermediate paths.

An example of the beam search algorithm can be found in Figure 5. Instead of greedily predicting the most probable label, the beam search algorithm finds the top-$N$ most probable prediction paths as *intermediate paths* $\mathcal{S}(t)$ at each time step $t$.

$$\mathcal{S}(t) = \{P_1(t), P_2(t), \cdots, P_N(t)\} \qquad (7)$$

At time step $t + 1$, we add $N$ most probable labels to each intermediate path $P_i(t)$ to get a total of $N \times N$ paths. The $N$ prediction paths with highest probability among these paths constitute the *intermediate paths* for time step $t + 1$. The prediction paths ending with the END sign are added to the *candidate path* set $\mathcal{C}$. The termination condition of the beam search is that the probability of the current intermediate paths is smaller than that of all the candidate paths. It indicates that we cannot find any more candidate paths with greater probability.

### 3.4. Training

Learning CNN-RNN models can be achieved by using the cross-entropy loss on the softmax normalization of score softmax($s(t)$) and employing back-propagation through time algorithm. In order to avoid the gradient vanishing/exploding issues, we apply the rmsprop optimization algorithm [33]. Although it is possible to fine-tune the convolutional neural network in our architecture, we keep the convolutional neural network unchanged in our implementation for simplicity.

One important issue of training multi-label CNN-RNN models is to determine the orders of the labels. In the experiments of this paper, the label orders during training are determined according to their occurrence frequencies in the training data. More frequent labels appear earlier than the less frequent ones, which corresponds to the intuition that easier objects should be predicted first to help predict more difficult objects. We explored learning label orders by iteratively finding the easiest prediction ordering and order ensembles as proposed in [28] or simply using fixed random order, but they do not have notable effects on the performance. We also attempted to randomly permute the label orders in each mini-batch, but it makes the training very difficult to converge.

## 4. Experiments

In our experiments, the CNN module uses the 16 layers VGG network [29] pretrained on ImageNet 2012 classification challenge dataset [5] using Caffe deep learning framework [16]. The dimensions of the label embedding and of LSTM RNN layer are 64 and 512, respectively. We employ weight decay rate 0.0001, momentum rate 0.9, and dropout [4] rate 0.5 for all the projection layers.

We evaluate the proposed method on three benchmark multi-label classification datasets: NUS-WIDE, Microsoft COCO, and VOC PASCAL 2007 datasets. The evaluation demonstrates that the proposed method achieves superior performance to state-of-the-art methods. We also qualitatively show that the proposed method learns a joint label/image embedding and it focuses its attention in different image regions during the sequential prediction.

### 4.1. Evaluation Metric

The precision and recall of the generated labels are employed as evaluation metrics. For each image, we generate $k$[1] highest ranked labels and compare the generated labels to the ground truth labels. The precision is the number of correctly annotated labels divided by the number of generated labels; the recall is the number of correctly annotated labels divided by the number of ground-truth labels.

We also compute the per-class and overall precision (C-P and O-P) and recall scores (C-R and O-R), where the average is taken over all classes and all testing examples, respectively. The F1 (C-F1 and O-F1) score is the geometrical average of the precision and recall scores. We also compute the mean average precision (MAP)@N measure [34].

### 4.2. NUS-WIDE

NUS-WIDE dataset [2] is a web image dataset that contains 269,648 images and 5018 tags from Flickr. There are a total of 1000 tags after removing noisy and rare tags. These images are further manually annotated into 81 concepts by human annotators. An example of the annotations and predictions for both of label set is shown in the left side of Fig. 6. The quality of 81-tag annotations is relatively high, while the 1000-tag annotations are very noisy. In Fig. 6, we can find some tags with duplicate semantics, such as "cloud" and "clouds", some completely wrong tags, such as "photographer", and some tags that are too general to have specific meanings, such as "beautiful".

---
[1] For RNN model, we set the minimum prediction length during beam search to ensure that at least $k$ labels are predicted.

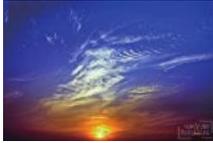

| | |
|---|---|
| Tags80: clouds, sun, sunset | |
| Tags1k: blue, clouds, sun, sunset, light, orange, photographer | Tags: car, person, luggage, backpack, umbrella, cup, truck |
| Prediction 80: clouds, sky, sun, sunset | |
| Prediction 1k: nature, sky, blue, clouds, red, sunset, yellow, sun, beatiful, sunrise, cloud | Predictions: person, car, truck, backpack |

Figure 6. One example image from NUS-WIDE (left) and MS-COCO(right) datasets, the ground-truth annotations and our model's predictions.

| Method | C-P | P-R | C-F1 | O-P | O-R | O-F1 | MAP@10 |
|---|---|---|---|---|---|---|---|
| Metric Learning [19] | - | - | - | - | - | 21.3 | - |
| Multi-edge graph [23] | - | - | - | 35.0 | 37.0 | 36.0 | - |
| KNN [2] | 32.6 | 19.3 | 24.3 | 42.9 | 53.4 | 47.6 | - |
| Softmax | 31.7 | 31.2 | 31.4 | 47.8 | 59.5 | 53.0 | - |
| WARP [9] | 31.7 | **35.6** | 33.5 | 48.6 | 60.5 | 53.9 | - |
| Joint Embedding [38] | - | - | - | - | - | - | 40.3 |
| CNN-RNN | **40.5** | 30.4 | **34.7** | **49.9** | **61.7** | **55.2** | **56.1** |

Table 1. Comparisons on NUS-WIDE Dataset on 81 concepts for $k = 3$.

We first evaluate the proposed method on less noisy 81 concepts labels. We compare the proposed method with state-of-the-art methods including K nearest neighbor search [2], softmax prediction, WARP method [9], metric learning [19], and joint embedding [38] in Table 1. Since there is less noise in 81 concepts labels, all methods achieve fairly good performance. Although we do not fine-tune our convolutional neural network image representation, the proposed RNN framework outperforms the state-of-the-art methods. In particular, we find the CNN-RNN framework achieves 8% higher precision, because it is capable of exploiting the label correlation to filter out the labels that can not possibly exist together.

We also compare RNN model with softmax, DSLR [22], and WARP models on the more challenging 1000-tag label set in Table 2. The prediction accuracy of all the methods are very low, because the labels on this dataset are very noisy, but the proposed method still outperforms all the baseline methods. We find the proposed method cannot distinguish gender-related labels such as "actor" and "actress", because our convolutional neural network is trained on ImageNet, which does not have the annotation for this task. More multi-label prediction examples can be found in the supplemental materials.

| Method | C-P | P-R | C-F1 | O-P | O-R | O-F1 | MAP@10 |
|---|---|---|---|---|---|---|---|
| Softmax | 14.2 | **18.6** | 16.1 | 17.1 | 28.8 | 21.5 | 24.3 |
| DLSR [22] | - | - | - | 20.0 | 25.0 | 22.4 | - |
| WARP | 14.5 | 15.9 | 15.2 | 18.3 | 30.8 | 22.9 | 24.8 |
| CNN-RNN | **19.2** | 15.3 | **17.1** | **18.5** | **31.2** | **23.3** | **26.6** |

Table 2. Comparisons on NUS-WIDE Dataset on 1000 tags for $k = 10$.

| Method | C-P | P-R | C-F1 | O-P | O-R | O-F1 | MAP@10 |
|---|---|---|---|---|---|---|---|
| Softmax | 59.0 | 57.0 | 58.0 | 60.2 | 62.1 | 61.1 | 47.4 |
| WARP | 59.3 | 52.5 | 55.7 | 59.8 | 61.4 | 60.7 | 49.2 |
| Binary cross-entropy | 59.3 | **58.6** | 58.9 | 61.7 | 65.0 | 63.3 | - |
| No RNN | 65.3 | 54.5 | 59.3 | 68.5 | 61.3 | 65.7 | 57.2 |
| CNN-RNN | **66.0** | 55.6 | **60.4** | **69.2** | **66.4** | **67.8** | **61.2** |

Table 3. Comparisons on MS-COCO Dataset for $k = 3$.

### 4.3. Microsoft COCO

Microsoft COCO (MS-COCO) dataset [21] is an image recognition, segmentation, and captioning dataset. It contains 123 thousand images of 80 objects types with per-instance segmentation labels. Among those images, 82783 images are utilized as training data, and 40504 images are employed as testing data. We utilize the object annotations as the labels. An example of the annotations and predictions of the MS-COCO dataset is shown in the right of Fig. 6. An interesting property of the MS-COCO dataset is that most images in this dataset contain multiple objects, and these objects usually have strong co-occurrence dependencies. For example, "baseball glove" and "sport ball" have high co-occurrence probability, while "zebra" and "cat" never appear together.

We compare the softmax, multi-label binary cross entropy, and WARP [9] models with the CNN-RNN model in Table 3. Since the number of the objects per image varies considerably in this dataset, we do not set the minimum length of the prediction path during beam search. It can be observed that the proposed method achieves much better performance both in terms of overall precision and recall. It has a slightly lower per-class recall because it may output less than $k$ labels for an image and it usually chooses not to predict the small objects that have little co-occurrence dependencies with other larger objects. We also replace the recurrent layer with a linear embedding layer in the proposed architecture and evaluate the performance. We find that removing the recurrent layer significantly affects the recall of the multi-label classification.

The per-class precision and recall of the proposed frame-work is shown in Fig. 7. We find the proposed framework performs very well on large objects, such as "person", "zebra", and "stop sign", and the objects with high dependencies with other objects or the scene, such as "sports bar" and "baseball glove". It performs very poorly on small objects with little dependencies with other objects, such as "toaster" and "hair drier", because the global convolutional neural network image features have limited discriminative ability to recognize small objects. There is zero prediction for "toaster" and "hair drier", resulting in zero precision and recall scores for these two labels.

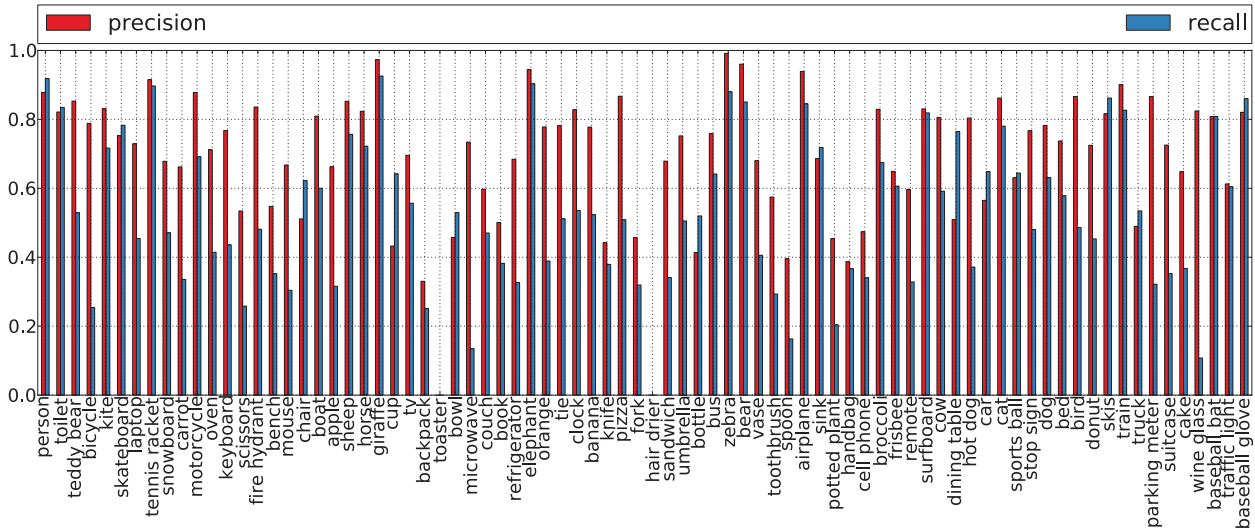

Figure 7. The per-class precision and recall of the RNN model on MS-COCO dataset.

| | plane | bike | bird | boat | bottle | bus | car | cat | chair | cow | table | dog | horse | motor | person | plant | sheep | sofa | train | tv | mAP |
|---|---|---|---|---|---|---|---|---|---|---|---|---|---|---|---|---|---|---|---|---|---|
| INRIA [14] | 77.2 | 69.3 | 56.2 | 66.6 | 45.5 | 68.1 | 83.4 | 53.6 | 58.3 | 51.1 | 62.2 | 45.2 | 78.4 | 69.7 | 86.1 | 52.4 | 54.4 | 54.3 | 75.8 | 62.1 | 63.5 |
| CNN-SVM [27] | 88.5 | 81.0 | 83.5 | 82.0 | 42.0 | 72.5 | 85.3 | 81.6 | 59.9 | 58.5 | 66.5 | 77.8 | 81.8 | 78.8 | 90.2 | 54.8 | 71.1 | 62.6 | 87.4 | 71.8 | 73.9 |
| I-FT [36] | 91.4 | 84.7 | 87.5 | 81.8 | 40.2 | 73.0 | 86.4 | 84.8 | 51.8 | 63.9 | 67.9 | 82.7 | 84.0 | 76.9 | 90.4 | 51.5 | 79.9 | 54.1 | 89.5 | 65.8 | 74.4 |
| HCP-1000C [36] | 95.1 | **90.1** | 92.8 | 89.9 | 51.5 | 80.0 | **91.7** | 91.6 | 57.7 | 77.8 | **70.9** | 89.3 | 89.3 | **85.2** | 93.0 | **64.0** | 85.7 | 62.7 | 94.4 | 78.3 | 81.5 |
| CNN-RNN | **96.7** | 83.1 | **94.2** | 92.8 | **61.2** | **82.1** | 89.1 | **94.2** | **64.2** | **83.6** | 70.0 | **92.4** | **91.7** | 84.2 | **93.7** | 59.8 | **93.2** | **75.3** | **99.7** | **78.6** | **84.0** |

Table 4. Classification results (AP in %) comparison on PASCAL VOC 2007 dataset.

The relationship between recall and bounding box area on Microsoft COCO dataset is shown in Fig. 8. We can observe that the recall is the generally higher as the object is larger, unless the object is so large that it almost fill the whole image, where some important information might lose during the image cropping process in CNN.

### 4.4. PASCAL VOC 2007

PASCAL Visual Object Classes Challenge (VOC) datasets [6] are widely used as the benchmark for multilabel classification. VOC 2007 dataset contains 9963 images divided into *train*, *val* and *test* subsets. We conduct our experiments on *trainval/test* splits (5011/4952 images). The evaluation is Average Precision (AP) and mean of AP

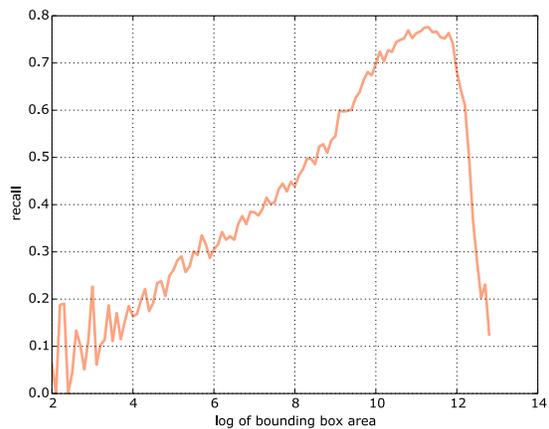

Figure 8. The relationship between recall and bounding box area on Microsoft COCO dataset.

(mAP).

The comparison to state-of-the-art methods is shown in Table 4. INRIA [14] is based on the transitional feature extraction-coding-pooling pipeline. CNN-SVM [27] directly applies SVM on the OverFeat features pre-trained on ImageNet. I-FT [36] employs the squared loss function on the shared CNN features on PASCAL VOC for multi-label classification. HCP-1000C [36] employs region proposal information to fine-tune the CNN features pre-trained on ImageNet 1000 dataset, and ac hives better performance than the methods that do not exploit region information. The proposed CNN-RNN method outperforms the I-FT method by a large margin, and it also performs better than HCP-1000C method, although the RNN method does not take the region proposal information into account.

Table 5. Nearest neighbors for label embeddings of 1k labels of NUS-WIDE and MS-COCO datasets

| Label | Nearest Neighbors |
|---|---|
| glacier | arctic, norway, volcano, tundra, lakes |
| sky | nature, blue, clouds, landscape, bravo |
| sunset | sun, landscape, light, bravo, yellow |
| rail | railway, track, locomotive, tracks, steam |
| cat | dog, bear, bird, hair drier, toaster |
| cow | horse, sheep, bear, zebra, elephant |

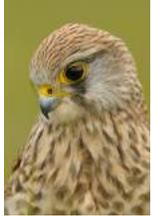

Figure 9. Nearest neighbor labels and top classification predictions by softmax model for three query images of 1000 label set on NUS-WIDE dataset.

### 4.5. Label embedding

In addition to being able to generate multiple labels, the CNN-RNN model also effectively learns a joint label/image embedding. The nearest neighbors of the labels in the embedding space for NUS-WIDE and MS-COCO, are shown in Table 5. We can see that a label is highly semantically related to its nearest-neighbor labels.

Fig. 9 shows the nearest neighbor labels for images on NUS-WIDE 1000-tag dataset computed according to label embedding $w_k$ and image embedding $U_I^x I$. In the joint embedding space, an image and its nearest neighbor labels are semantically relevant. Moreover, we find that compared to the top-ranked labels predicted by classification model, the nearest neighbor labels are usually more fine-grained. For example, the nearest neighbor labels "hawk" and "glacier" are more fine-grained than "bird" and "landscape".

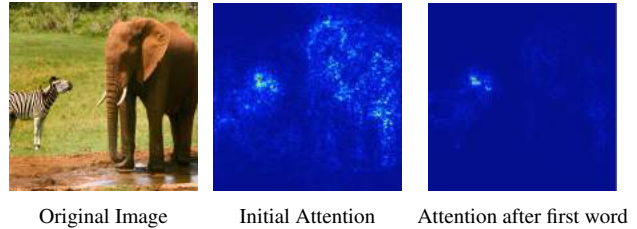

Original Image    Initial Attention    Attention after first word

Figure 10. The attentional visualization for the RNN multi-label framework. This image has two ground-truth labels: "elephant" and "zebra". The bottom-left image shows the framework's attention in the beginning, and the bottom-right image shows its attention after predicting "elephant".

### 4.6. Attention Visualization

It is interesting to investigate how the CNN-RNN framework's attention changes when predicting different labels. We visualize the attention of the RNN multi-label model with Deconvolutional networks [40] in Fig. 10. Given an input image, the attention of the RNN multilabel model at each time step is the average of the synthesized image of all the label nodes at the softmax layer using Deconvolutional network. The ground truth labels of this image are "elephant" and "zebra". (Notice that the visualization of attention does not utilize the ground truth labels) At the beginning, the attention visualization shows that the model looks over the whole image and predicts "elephant". After predicting "elephant", the model shifts its attention to the regions of zebra and predicts "zebra". The visualization shows that although the RNN framework does not learn an explicit attentional model, it manages to steer its attention to different image regions when classifying different objects.

## 5. Conclusion and Future Work

We propose a unified CNN-RNN framework for multi-label image classification. The proposed framework combines the advantages of the joint image/label embedding and label co-occurrence models by employing CNN and RNN to model the label co-occurrence dependency in a joint image/label embedding space. Experimental results on several benchmark datasets demonstrate that the proposed approach achieves superior performance to the state-of-the-art methods.

The attention visualization shows that the proposed model can steer its attention to different image regions when predicting different labels. However, predicting small objects is still challenging due to the limited discriminativeness of the global visual features. It is an interesting direction to not only predict the labels, but also predict the segmentation of the objects by constructing an explicit attention model. We will investigate that in our future work.